\documentclass[]{llncs}

\usepackage{graphicx}
\usepackage[disable]{todonotes}

\begin{document}

{\let\thefootnote\relax\footnotetext{Copyright \textcopyright\ 2019 for this paper by its authors. Use permitted under Creative Commons License Attribution 4.0 International (CC BY 4.0). CLEF 2019, 9-12 September 2019, Lugano, Switzerland.}}

\title{Overview of CLEF 2019 Lab ProtestNews: Extracting Protests from News in a Cross-context Setting}
\author{Ali H\"{u}rriyeto\u{g}lu \and
Erdem Y\"{o}r\"{u}k \and
Deniz Y\"{u}ret \and
\c{C}a\u{g}r{\i} Yoltar \and
Burak G\"{u}rel \and
F{\i}rat Duru\c{s}an \and 
Osman Mutlu \and
Arda Akdemir}
\authorrunning{H\"{u}rriyetoglu et al.}
\institute{Koc University, Istanbul 34450, Turkey \\
\email{\{ahurriyetoglu,eryoruk,dyuret,cyoltar,bgurel,fdurusan,omutlu, aakdemir\}@ku.edu.tr}\\
\url{http://www.ku.edu.tr}}
\maketitle              % typeset the header of the contribution
\begin{abstract}

We present an overview of the CLEF-2019 Lab ProtestNews on Extracting Protests from News in the context of generalizable natural language processing. The lab consists of document, sentence, and token level information classification and extraction tasks that were referred as task 1, task 2, and task 3 respectively in the scope of this lab. The tasks required the participants to identify protest relevant information from English local news at one or more aforementioned levels in a cross-context setting, which is cross-country in the scope of this lab. The training and development data were collected from India and test data was collected from India and China. The lab attracted 58 teams to participate in the lab. 12 and 9 of these teams submitted results and working notes respectively. We have observed neural networks yield the best results and the performance drops significantly for majority of the submissions in the cross-country setting, which is China. 

\end{abstract}
\section{Introduction}

We describe a realization of our task set proposal~\cite{Hurriyetoglu+19} in the scope of CLEF-2019 Lab ProtestNews.\footnote{\url{http://clef2019.clef-initiative.eu/}}$^{,}$\footnote{\url{https://emw.ku.edu.tr/clef-protestnews-2019/}} The task set aims at facilitating development of generalizable natural language processing (NLP) tools that are robust in a cross-context setting, which is cross-country in this lab. Since the performance of NLP tools significantly drop in a context different from the one they are created and validated \cite{Akdemir+2018,Ettinger+17,Soboroff+18}, measuring and improving state-of-the-art NLP tool development methodology is the primary aim of our efforts. 

Comparative social and political science studies facilitate protest information to analyze cross-country similarities, differences, and effect of these actions. Therefore our lab focuses on classifying and extracting protest event information in English local news articles from India and China. We believe our efforts will contribute to enhance the methodologies applied to collect data for these studies. This need was motivated based on the recent results that shows NLP tools, those of text classification and information extraction, have not been satisfactory against the requirements of longer time coverage and working on data from multiple countries \cite{Wang+16,Hammond+14}.

This first iteration of our lab attracted 58 teams from all around the world. 12 of these teams submitted their results to one or more tasks on the CodaLab page of the lab.\footnote{\url{https://competitions.codalab.org/competitions/22349}} 9 teams described their approach in terms of a working note.

We introduce the task set we tackle, the corpus we have been creating, and the evaluation methodology in Sections \ref{tasks}, \ref{data}, and \ref{evaluation} respectively. We report the results in Section \ref{results} and conclude our report in Section \ref{conclusion}. %We discuss future directions of our work in Section \ref{future-work}.

\section{Task Set}
\label{tasks}
\begin{figure}
\centering
\includegraphics[width=0.8\textwidth]{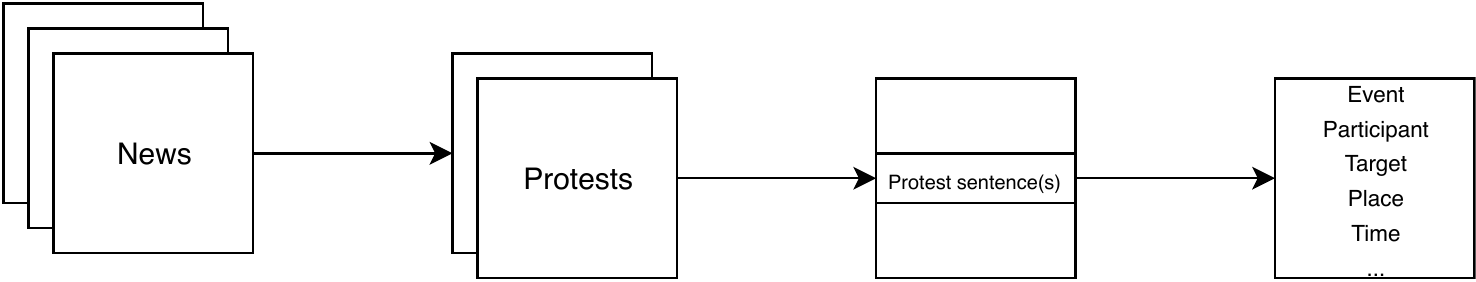}
\caption {The lab consists of a) {\em Task 1}: Document classification, b) {\em Task 2}: Event sentence detection, and c) {\em Task 3}: Event extraction.} \label{fig:tasksfigure}
\end{figure}

The lab consists of the tasks document classification, event sentence detection and event extraction, which are referred as task 1, task 2, and task 3 respectively, as demonstrated in Figure~\ref{fig:tasksfigure}. The document classification task, which is task 1, requires predicting whether a news article report at least one protest event that has happened or is happening. It is a binary classification task that require to predict whether a news article label should be 1 (positive) or 0 (negative). The sentences that contain any event trigger should be identified in task 2, which is event sentence detection task. Sentence labels are 0 and 1 as well. This task could be handled either as classification or extraction task as we provide order of the sentences in their respective articles. Finally, the event triggers and event information, which are place, facility, time, organizer, participant, and target, should be extracted in task 3. This order of tasks provides a controlled setting that enables error analysis and optimization possibility during annotation and tool development efforts. Moreover, this design enable analyzing steps of the analysis that contributes to explainability of the automated tool predictions.

\section{Data}
\label{data}

We provide the number of instances for each task in Table~\ref{tab:datasize} in terms of training, development, test 1, and test 2 data. The training and development data was collected from online local English news from India. Test 1 and test 2 data refer to data from India and China respectively.

\begin{table}
\centering
\caption{Number of instances for each task}\label{tab:datasize}
\setlength{\tabcolsep}{10pt} 
\begin{tabular}{|l|c|c|c|c|c|}
\hline
  & Training & Development & Test 1 (India) & Test 2 (China)  \\
\hline
Task 1 & 3,429 & 456 & 686 & 1,800 \\
Task 2 & 5,884 & 662 & 1,106 & 1,234 \\
Task 3 & 250 & 36 & 80 & 39 \\
\hline
\end{tabular}
\end{table}

A sample from task 1 contains the news article's text, its URL and its label that is assigned by the annotation team. For task 2: sentences, their labels, their order in the article's text they belong, and the URL of their article are available in samples. The release format of the data for task 2 enable participants to treat this task either as classification of individual sentences or extracting event relevant sentences from a document.

\begin{figure}[h]
    \centering
    \includegraphics[width=0.9\textwidth]{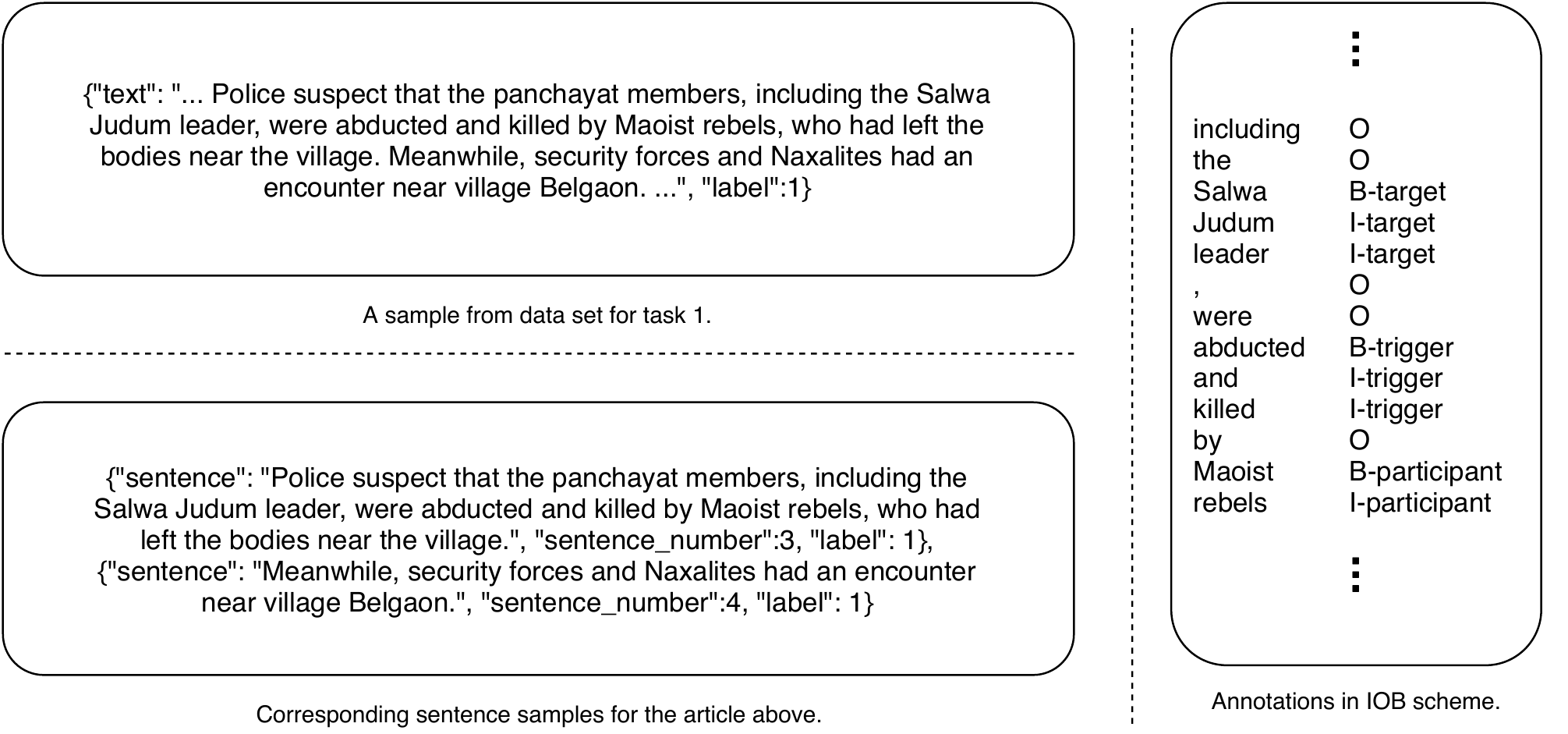}
    \caption{Data samples for task 1, task 2, and task 3}
    \label{fig:data-examples}
\end{figure}

The data for task 3 consists of snippets that contain one or more event trigger that refer to the same event. Multiple sentences may occur in a snippet in case these sentences refer to the same event.\footnote{Snippets we share contain information about only a single event.} The tokens in these snippets are annotated using IOB, inside, outside, beginning, scheme. The examples of data is provided in Figure~\ref{fig:data-examples}. 

There is not any overlap of news articles across tasks. This separation was required in order to avoid any misuse of data from one task to infer the labels for another task without any effort.

\subsection{Distribution}

We distributed the data set in a way that does not violate copyright of the news sources. This involves only sharing information that is needed to reproduce the corpus from the source for task 1 and task 2 and only relevant snippets for task 3. We released a Docker image that contains the toolbox\footnote{\url{https://github.com/emerging-welfare/ProtestNews-2019}} required to reproduce the news articles on the computer of a participant. The toolbox generates a log of the process that reproduce the data set and we have requested these log files from the participants. The toolbox is a pipeline that scrapes HTMLs, converts HTMLs to text and finally performs specific filling operation for each of task 1 and task 2. To the best of our knowledge, the toolbox succeeded in enabling participants create the data set on their computers. Only one participant from Iran was not able to download the news articles due to restrictions to access online content that are specific to his geolocation.

\section{Evaluation Setting}
\label{evaluation}

We use macro averaged F1 due to class imbalance present in our data for evaluating the task 1 and task 2. The event extraction task, which is task 3, was evaluated on the average F1 score of all information types that was based on the ratio of the full match between the prediction and the annotations in the test sets, using a python implementation\footnote{https://github.com/sighsmile/conlleval} of CoNLL 2003 shared task \cite{sang2003introduction} evaluation script.

We performed two levels of evaluation that were on data from the source country (Test 1) and from target country (Test 2). The participants were informed only about labels of the training and development data from the source country. They did not see labels of any test set. The number of allowed submissions and the period the participants can submit their predictions for Test 1 and Test 2 was determined in a way that restrict the possibility of over-fitting on test data. We limited the number of submission and the submission period in order to make sure the participants do not overfit to the test data based. 

We applied three cycles of evaluation. Participants could submit unlimited number of results without being able to see their score. Their last submitted results' scores were announced at the end of each evaluation cycle. First and second cycles aimed at providing feedback to the participants. The third and final cycle was the deadline for submitting results.

Finally, we provided a baseline submission for task 1 and task 2 in order to guide the participants. This baseline was based on predictions of the the best scoring machine learning model among Support Vector Machines, Naive Bayes, Rocchio classifier, Ridge Classifier, Perceptron, Passive Aggressive Classifier, Random Forest, K Nearest Neighbors, and Elastic Net on development set. The best scoring model was a linear support vector machines classifier that was trained using stochastic gradient descent.

\section{Results}
\label{results}

We facilitated CodaLab platform for managing the submissions and maintain a leaderboard.\footnote{\url{https://competitions.codalab.org/competitions/22349\#results}} The leaderboard for task 1 and task 2 is presented in Table~\ref{tab:resultsTask1task2}. The column names have the following format Test $<$task number$>$-$<$test number$>$, e.g. Test 1-1 stands for the Test 1 of task 1. The results for ProtestLab\_Baseline is the aforementioned baseline that was submitted by us.

\begin{table}
\centering
\caption{Results that are ranked based on average F1 scores of Test 1 and Test 2 for task 1 and task 2}\label{tab:resultsTask1task2}
\setlength{\tabcolsep}{10pt} 
\begin{tabular}{|l|c|c|c|c|c|c|}
\hline
  & Test 1-1 & Test 1-2 & Test 2-1  & Test 2-2 & Avg \\
\hline
ASafaya&.81&.63 &\textbf{.70}&.60&.69\\
PrettyCrocodile &.79&.60&.65&\textbf{.64}&.67\\
LevelUp Research &\textbf{.83}&\textbf{.65}&.66&.45&.65\\
Provos\_RUG &.80&.59&.63&.55&.64\\
GS &.78 &.56&.64&.58&.64\\
Be-LISI &.76&.50&.58&.30&.54\\
ProtestLab\_Baseline &.83&.49&.58&.20&.52\\
CIC-NLP &.59&.50&.52&.34&.49\\
SSNCSE1 &.38&.15&.56&.35&.36\\
iAmirSoltani &.69&.36&-&-&.26\\
Sayeed Salam &.55&.28&-&-&.20\\
SEDA lab &.58&-&.15&.02&.19 \\
\hline
\end{tabular}
\end{table}

The summary of the approaches and results of each team that participated in task 1 and or task 2 are provided below.\footnote{We have not received details of the submissions from CIC-NLP, iAmirSoltani, and Sayeed Salam. The details of other approaches can be found in the respective working notes that were published in proceedings of CLEF 2019 Lab ProtestNews.}

\begin{description}
\item [ASafaya (Sakarya University)] submitted the best results for task 2 and for average of task 1 and task 2 using Bidirectional Gated Recurrrent Unit (GRU) based model. Although this model perform the best on average, the performance of this model drops across the context significantly.
\item [PrettyCrocodile (National Research University HSE)] submitted the second best average results that were predicted using Embeddings from Language Models (ELMo). The performance of the model is comparable in the cross-context setting for task 2.
\item [LevelUp Research (University of North Carolina at Charlotte)] has applied multi-task learning based on LSTM units using word embeddings from a pre-trained FastText model. This method yielded the best results for task 1.
\item [Provos\_RUG (University of Groningen)] has implemented a feature based stacked ensemble model based on FastText embeddings and a set of different basic Logistic Regression classifiers that enabled their predictions to rank fourth among the participating teams.
\item [GS (University of Bremen)] has stacked the word embeddings such as GloVe and FastText together with the contextualized embeddings generated from Flair language models (LM). This approach was ranked fourth in general and third for task 2. 
\item [Be-LISI (Université de Carthage)] combined the logistic regression with
linguistic processing and expansion of the text with related terms using word embedding similarity. This approach marked a significant difference in terms of overall performance, which is the drop from .64 to .54, in comparison to higher ranked submissions.
\item [SSNCSE1 (Sri Sivasubramaniya College of Engineering)] reported results of their bi-directional LSTM that applies Bahdanau, Normed-Bahdanau, Luong, and Scaled-Luong attentions. The submission that uses Bahdanau attention yielded the results reported in Table~\ref{tab:resultsTask1task2}.
\item [SEDA lab (University of Exeter)] applied support vector machines and XGBoost classifiers that are combined with various word embedding approaches. Results of this submission showed promising performance in terms of precision on both document and sentence classification tasks.
\end{description}

% 	Task 1 India	Task 1 China	Task 2 India	Task 2 China	Average
% alisafaya	0.81	0.63	0.70	0.60	0.69
% makslizok	0.79	0.60	0.65	0.64	0.67
% mil24havoc	0.83	0.65	0.66	0.45	0.65
% tcaselli	0.80	0.59	0.63	0.55	0.64
% gabriellasky	0.78	0.56	0.64	0.58	0.64
% chedi	0.76	0.50	0.58	0.30	0.54
% ProtestLab_Baseline	0.83	0.49	0.58	0.20	0.52
% saroyehun	0.59	0.50	0.52	0.34	0.49
% Abishek_Shyamsunder	0.38	0.15	0.56	0.35	0.36
% AmirSoltani	0.69	0.36	-	-	0.26
% sayeedsalam	0.55	0.28	-	-	0.20
% AnaisO	0.58	-	0.15	0.02	0.19

We analyze task 3 results separate from task 1 and task 2 as it differs from them. The F1 scores for task 3 are presented in Table~\ref{tab:resultsTask1task2}. 

\begin{table}
\centering
\caption{Results that are ranked based on average score of Test 1 and Test 2 for Task 3}\label{tab:resultsTask1task3}
\setlength{\tabcolsep}{10pt} 
\begin{tabular}{|l|c|c|c|c|}
\hline
  & Test 3-1 & Test 3-2 & Avg \\
  \hline
GS	& .604 & .532 & .568\\
DeepNEAT &	.601 & .513 & .557\\
Provos\_RUG &	.600 & .456 & .528\\
PrettyCrocodile &	.524 & .425 & .474\\
LevelUp Research	& .516 & .394 & .455\\
\hline
\end{tabular}
\end{table}

\begin{description}
\item [GS (University of Bremen)] submitted the best results for task 3 using a BiLSTM-CRF model incorporating pooled contextualized flair embeddings, and their model was the best in generalizing.
\item [DeepNEAT (FloodTags \& Radboud University)] compares the submitted ELMO+BiLSTM model to a traditional CRF and shows that the former is better and more generalizable.
\item [Provos\_RUG (University of Groningen)] divides the task 3 into two as event trigger detection task and event argument detection task using BiLSTM-CRF model with word embeddings, POS embeddings, and character-level embeddings for both subtasks. He further extends the features for latter subtask with learned embeddings for dependency relations and event triggers.
\item [PrettyCrocodile (National Research University HSE)] makes use of ELMO embeddings with different architectures, achieving her best score for task 3 using a BiLSTM.
\item [LevelUp Research (University of North Carolina at Charlotte)] implemented a multi-task neural model that require a time-ordered sequence of word vectors representative of a document or sentence. The LSTM layer has been replaced by a layer of bidirectional gated recurrent units (GRU).
\end{description}

\section{Conclusion}
\label{conclusion}

The results show how text classification and information extraction tool performances drops between  two contexts. The scores on data from the target country are significantly lower than on data from the source country. Only the PrettyCrocodile team performed comparatively well across contexts for task 2. Although it is not the best scoring system for neither task 1 nor task 2, PrettyCrocodile team's approach show some promise toward tackling the generalizability of NLP tools.

The generalization of automated tools is an issue that has recently attracted much attention.\footnote{\url{https://sites.google.com/view/icml2019-generalization/cfp}} However, as we have determined in our lab, generalizability is still a challenge for state-of-the-art methodology.  Consequently, we will continue our efforts by repeating this practice and extending the data and will be adding data from new countries and languages to our setting. The next iteration will run in the scope of the Workshop on Challenges and Opportunities in Automated Coding of COntentious Political Events (Cope 2019) at European Symposium Series on Societal Challenges in Computational Social Science (Euro CSS 2019).\footnote{\url{https://competitions.codalab.org/competitions/22842}}$^{,}$\footnote{\url{https://emw.ku.edu.tr/?event=challenges-and-opportunities-in-automated-coding-of-contentious-political-events\&event\_date=2019-09-02}}$^{,}$\footnote{http://symposium.computationalsocialscience.eu/2019/}

\todo[inline]{- annoation manual'lerden örnekler verebiliriz.
- annotation metadolojisini anlatabiliriz
- future work'u biraz daha genişletebiliriz. Aynı deadline'lı CLEF 2020 proposal'ı var. Ektekini düzenliyor olacağım, bu paper'a da aktarabileceğim şeyler olabilir.
- field'in yapması gerekenleri biraz daha açabiliriz: event tanımları ve buna göre veri üretme konusunda collaborative efort gibi
- task 3 (event extraction) için detaylı bir hata analizi yapabiliriz}

\section*{Acknowledgments}
This work is funded by the European Research Council (ERC) Starting Grant 714868 awarded to Dr. Erdem Y\"{o}r\"{u}k for his project Emerging Welfare. We are grateful to our steering committee members for the CLEF 2019 lab Sophia Ananiadou, Antal van den Bosch, Kemal Oflazer, Arzucan \"{O}zg\"{u}r, Aline Villavicencio, and Hristo Tanev. Finally, we thank to Theresa Gessler and Peter Makarov for their contribution in organizing the CLEF lab by reviewing the annotation manuals and sharing their work with us respectively. 

\bibliographystyle{splncs04}
\bibliography{emwwp2}




% LNCS Homepage, \url{http://www.springer.com/lncs}. Last accessed 4
% \end{thebibliography}
\end{document}